\documentclass[a4paper]{esannV2}
\usepackage{times} %
\usepackage[utf8]{inputenc}

\usepackage{graphicx}
\graphicspath{{./plots/}}
\usepackage{afterpage}
\usepackage{amsmath} %
\usepackage{amssymb} %
\usepackage{mathtools}
\usepackage{bigints}

\usepackage{bm}
\usepackage{algorithm}
\usepackage[noend]{algpseudocode}
\usepackage{subcaption}
\usepackage{epstopdf}
\usepackage{stfloats}

\usepackage{soul}
\usepackage{tikz}
\usepackage{xcolor}
\usepackage{pgfplots}
\pgfplotsset{compat=1.14}
\usepgfplotslibrary{groupplots}
\usepackage{booktabs}
\usepgfplotslibrary{external} 
\usetikzlibrary{external} %
\makeatletter

\let\c@author\relax
\makeatother
\usepackage[backend=bibtex,sorting=none,natbib=true,url=false,doi=false]{biblatex}

\AtNextBibliography{\footnotesize}
\begin{document}
\title{Zero-shot and few-shot time series forecasting with ordinal regression recurrent neural networks}

\author{Bernardo~P\'erez Orozco$^{1,2}$ and Stephen J.~Roberts$^{1,2}$
\thanks{We acknowledge the support provided by CONACYT (CVU \#598304) and the UK Royal Academy of Engineering.}
\vspace{.3cm}\\
1- Information Engineering, University of Oxford
\vspace{.1cm}\\
2 - Mind Foundry Ltd. $\quad$
}

\maketitle

\begin{abstract}
Recurrent neural networks (RNNs) are state-of-the-art in several sequential learning tasks, but they often require considerable amounts of data to generalise well. For many time series forecasting (TSF) tasks, only a few dozens of observations may be available at training time, which restricts use of this class of models. We propose a novel RNN-based model that directly addresses this problem by learning a shared feature embedding over the space of many quantised time series. We show how this enables our RNN framework to accurately and reliably forecast unseen time series, even when there is little to no training data available.
\end{abstract}
	\section{INTRODUCTION}
\label{sec:intro}
Deep and recurrent neural networks (RNNs) have achieved great success across a wide range of sequential learning tasks, such as machine translation, speech recognition and time series forecasting (TSF) \citep{Sutskever2014}. However, RNNs are ultra-parameterised models that overfit without sufficient data at hand, making them unfit to solve  tasks where few data are available. We propose a novel RNN approach to infer full predictive distributions for TSF tasks where there is little to no training data at hand, respectively referred to as few-shot and zero-shot learning. We achieve this by learning a shared feature representation across multiple forecasting tasks, then transferring this  knowledge to help predict unseen time series, even in observation poor environments. We show that our approach outperforms other (widely used) approaches across a range of metrics as well as providing a framework to visualise and interpret the embedding learnt by the neural network. 

\textbf{Related work.} There is a rich literature associated with transfer learning and few-shot learning, especially in the Computer Vision community, where pre-trained models such as VGGNet, ImageNet and AlexNet are regularly used as either transferrable feature extractors, followed by a simpler classifiers, or as initial priors before fine-tuning a model.

However, to the best of the authors' knowledge, the transfer learning literature for time series forecasting is significantly more scarce and often limited to evaluations solely based on the residuals of point forecasts. Full probabilistic forecasts are, however, desirable as this enables practitioners to incorporate predictive uncertainty in their decision-making processes. This is often achieved using Gaussian Processes (GPs) \citep{Rasmussen2006}, non-parametric probabilistic models that are known to generalise well even with little data.
	\section{METHODOLOGY}
\label{sec:methods}

Our strategy consists in inferring a joint Memory-endowed Ordinal Regression Deep Neural Network (MOrdReD) \citep{perez2018mordred} over a collection of (auxiliary) time series. This results in a cross-task feature embedding that we use as a basis to forecast unseen time series even with scarce data. We briefly summarise this model below, but refer the reader to \citep{perez2018mordred} for a more detailed explanation. 

\subsection{Memory-endowed Ordinal Regression Deep Neural Networks}\label{subsec:mordred}
MOrdReD is an ordinal regression approach to time series forecasting that has been shown to achieve long-term, reliable performance. In the ordinal setting, real-valued time series observations are firstly discretised into $M$ (normally equal-sized) subintervals $\mathcal{C} = \{\mathcal{C}_k\}_{k=1}^M$ and encoded using a $1$-of-$M$ scheme. Such discrete symbol sequences have been most efficiently dealt with in NLP tasks by deep recurrent neural networks and in particular sequence-to-sequence extensions of the Long Short-Term Memory (LSTM) model \citep{Sutskever2014}.

Sequence-to-sequence models fit together two LSTM layers, an encoder $f^{(\text{enc})}$ and a decoder $f^{(\text{dec})}$. In the first stage of the model's \textit{forward pass} the encoder sequentially scans $P$ measurements of the time series $\mathbf{X}^{(\tau-1)} = (\mathbf{x}_{\tau - P},\dots,\mathbf{x}_{\tau - 1}) $ and summarises them as temporal characteristics $\mathbf{h}^{(\text{dec})}_0, \mathbf{C}^{(\text{dec})}_0$. These are then fed as initial states to the LSTM decoder, which iteratively produces an output sequence. The $(t+1)$-th element of the output sequence, $\hat{\mathbf{x}}_{t+1}$, is then given by:
\begin{align*}
	(\mathbf{h}^{(\text{dec})}_0, \mathbf{C}^{(\text{dec})}_0) &= f^{(\text{enc})}(\mathbf{X}^{(\tau-1)}),\\
	\hat{\mathbf{x}}_{t+1} &= \text{Softmax}\left( f^{(\text{dec})}(\mathbf{h}^{(\text{dec})}_{t-1}, \mathbf{C}^{(\text{dec})}_{t-1}, \mathbf{x}_{t})\right).
\end{align*}
The softmax output $\hat{\mathbf{x}}_{t+1}$ can be interpreted as both a categorical distribution over the set of ordinal bins, $\mathcal{C} = \{\mathcal{C}_k\}_{k=1}^M$, at time $t+1$, and as a piece-wise uniform over the original time series range. Output sequences are computed by iteratively feeding back $\hat{\mathbf{x}}_{t+1}$ into the model and predictive posterior distributions can readily be approximated through Monte Carlo Dropout \citep{pmlr-v48-gal16}.

\subsection{A General Unified Model for time series forecasting}\label{subsec:gum}
Our GUM approach is based on inferring a shared representation space of time series. This shared embedding is learnt by fitting a joint MOrdReD model over a single compilation dataset $\mathcal{X}_{\text{aux}}$ of 20 different auxiliary time series which are assumed to possess sufficient observations for training. The datasets were drawn from a variety of application domains, and further detail is given in \url{https://github.com/bperezorozco/ordinal_tsf}. 

Crucially, our ordinal regression cross-task embedding is learnt over symbolic patterns that are invariant to the original scale of the time series, as all auxiliary time series are quantised independently, but with the same number of bins $M=150$. A variety of shapes and motifs from the auxiliary datasets are thus encoded in a feature space of ordinal bin motifs. We analyse this cross-task embedding in Section \ref{sub:embedding}. 

We note that the quantiser proposed in \citep{perez2018mordred} requires prior knowledge of the time series range. With scarce data, the observed range could change within short predictive horizons. We handle this with a simple linear heuristic that estimates the range boundaries within a predictive horizon $P_h$. Given the largest first-order finite difference of the time series, $\Delta_{\text{max}}$, our linear heuristic approximates the boundaries $\mathbf{\tilde{X}}_\text{min}, \mathbf{\tilde{X}}_\text{max} = \mathbf{X}_\text{min} -P_h\Delta_{\text{max}}, \mathbf{X}_\text{max} +P_h\Delta_{\text{max}}$. Future work for improving this estimation includes relating it to Lipschitz constant inference  \citep{Calliess2015BayesianLC}. 

	\section{EXPERIMENTS}
\label{sec:experiments}

\subsection{Task 1: Zero-shot forecasting}\label{sub:pred}
We first assess the model's performance at zero-shot time series forecasting. %

\noindent \textbf{Data.} We used two dataset collections in our experiments. The first was used to fit our GUM ordinal regression neural network, and contains 20 varied auxiliary datasets drawn from \citep{Fulcher20130048} with outputs evenly quantised over $M=150$ bins. A copy of our auxiliary datasets is provided in \url{https://github.com/bperezorozco/ordinal_tsf}. The second collection is our test data, which corresponds to the yearly-frequency segment of the M4 competition dataset \citep{makridakis2018m4}. We drew 150 excerpts that had no missing data (no interpolation needed) and that had at least 36 observations available.

\noindent \textbf{Task.} Every M4 time series was split into $T=21$ timesteps, assumed to be observed, and $P_h=15$ timesteps for prediction. For our baselines, each segment represents the training and test data, respectively; for GUM, the quantised 21-sample excerpt is fed as the encoder input sequence, before the decoder computes the forecast with horizon $P_h=15$. Importantly, no neural network parameters are updated whilst scanning the 21-sample sequence.

\noindent \textbf{Baselines.} We compared our GUM neural network against two popular models: Gaussian Process Regression (GPR) \citep{Rasmussen2006} and state-space AR($p$) models \citep{durbin2012time}. Detailed derivations of both methods can be found in the given references. In the case of GPR, we fitted two instances for every time series: one with the M\'atern 5/2 kernel and one with the Rational Quadratic kernel, both commonly found in the literature to model a wide range of physical phenomena. In the case of State-space AR($p$), we fitted two instances with lookback $p=3,4$.

\noindent \textbf{Metrics.} We evaluated models using three metrics: (1) the \textbf{Negative log-likelihood (NLL):} through this we assess how well  predictive densities capture the ground truth prediction; (2) \textbf{The root mean squared error (RMSE):} to measure the mean predictive accuracy of the models; and (3) \textbf{Quantile-Quantile (QQ) distance:} QQ plots are graphical tools used to assess how well predictive densities match empirical probabilities. We summarise how well calibrated predictive distributions are through the QQ distance, defined as the deviation between the identity and the obtained QQ plot. We refer the reader to \citep{perez2018mordred} for full details.

\noindent \textbf{Results.} We present our results in Table \ref{tab:res}. The Table is divided into two sections. On the left side, we show the percentage of datasets in which our GUM neural network outperforms each of our benchmarks for every metric. We show that across all metrics, our model consistently performs better than all baselines for the majority of cases. On the right side of Table \ref{tab:res}, we provide the average rank of our models. Ranks from 1 (best) to 3 (worst) were assigned for every metric, dataset and model and averaged across all M4 evaluation datasets. In the case of GPR and AR($p$), the instance with the best performance across kernels or model order $p$ was chosen. This metric shows that, on average, we expect GUM to outperform all other models; and additionally, when it does not, it can still be expected to come as a second competitor.
\begin{table}[t!]
\centering
\small
\begin{tabular}{l|rrrr|rrr|rrrr}
\toprule
{} & \multicolumn{4}{|c|}{\textbf{(\%) GUM outp.}} & \multicolumn{3}{|c|}{\textbf{Zero-shot rank}} & \multicolumn{4}{|c|}{\textbf{Few-shot rank}} \\
\midrule 
{} &   AR3 &       AR4 & M5/2 &   RQ &   GUM & GP & AR  & GUM & Mor & GP & AR \\
\midrule
\textbf{NLL}         &               65 &  70 &  59 &  62 &             \textbf{1.76}&  2.04 &  2.20 & \textbf{ 1.1}&3.1&3.2&2.6\\
\textbf{RMSE} &               64 &  68 &  54 &  58 &             \textbf{1.83} &  2.01 &  2.15 & \textbf{1} & 3.1 & 2.9 & 3 \\
\textbf{QQD}     &               74 &  75 &  67 &  67 &             \textbf{1.65} &  2.01 &  2.34 & \textbf{1.8} &2.3	& 3 &	2.9  \\
\bottomrule
\end{tabular}
\caption{\textbf{Left}: percentage of unseen evaluation datasets where GUM outperforms each of our baselines in the zero-shot setting.
\textbf{Centre}: average rank achieved by each model in the zero-shot setting. Ranks 1 (best) to 3 were assigned to each model for each dataset and metric.  \textbf{Right}: average rank achieved by each model in the few-shot setting. Ranks 1 (best) to 4 were assigned to each model for each dataset and metric. 
}
\label{tab:res}
\end{table}

\subsection{Cross-task feature embedding analysis}\label{sub:embedding}
\begin{figure}[t!]
    \centering
 	\includegraphics[width=\textwidth]{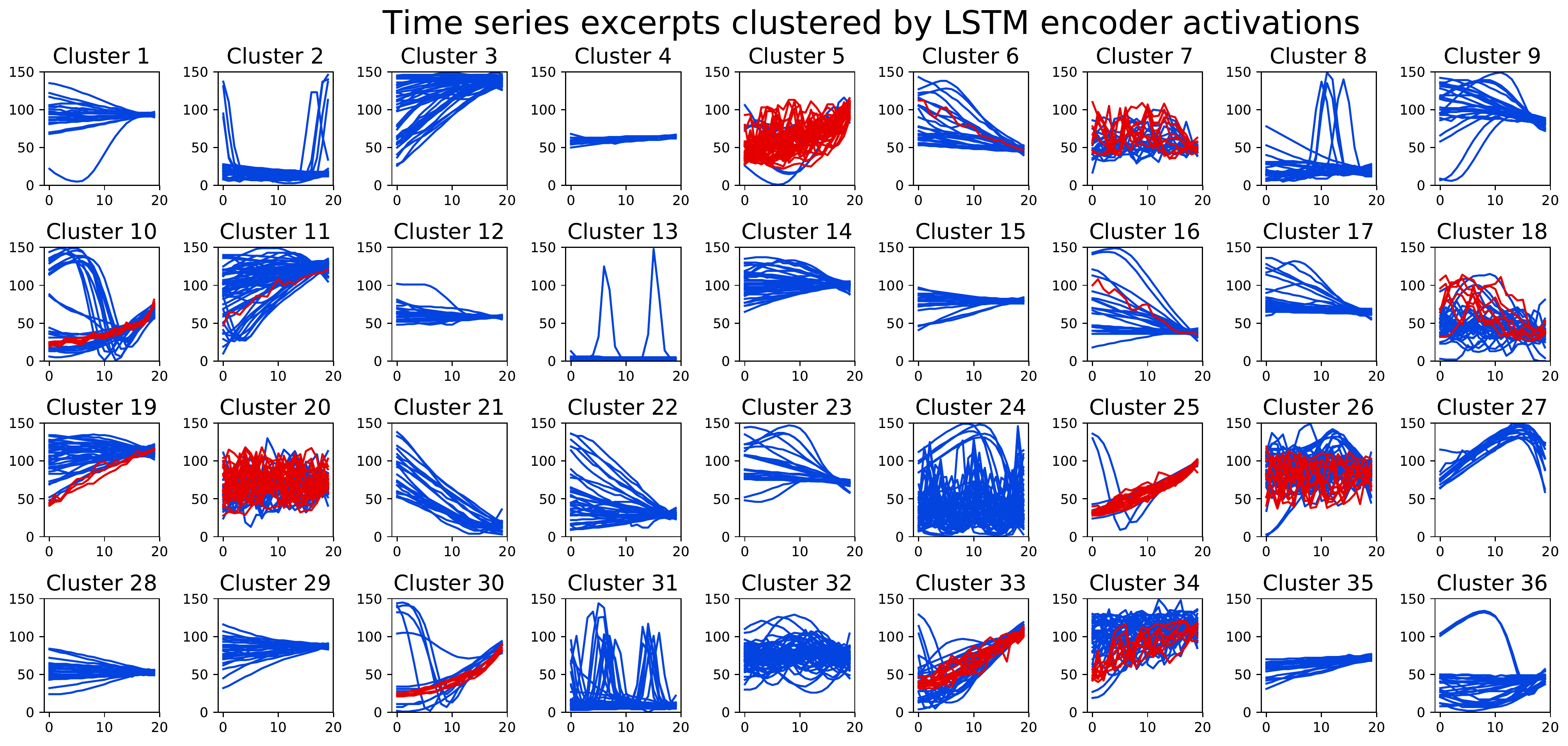}\caption{Sample excerpt groups resulting from clustering the LSTM encoder activations over both auxiliary (in blue) and M4 unseen time series (in red). Frames in the same cluster trigger similar LSTM activations and thus produce similar forecasts. }
 	\label{fig:frames}
 \end{figure}

\begin{figure}[t!]
    \centering
 	\includegraphics[width=0.85\textwidth]{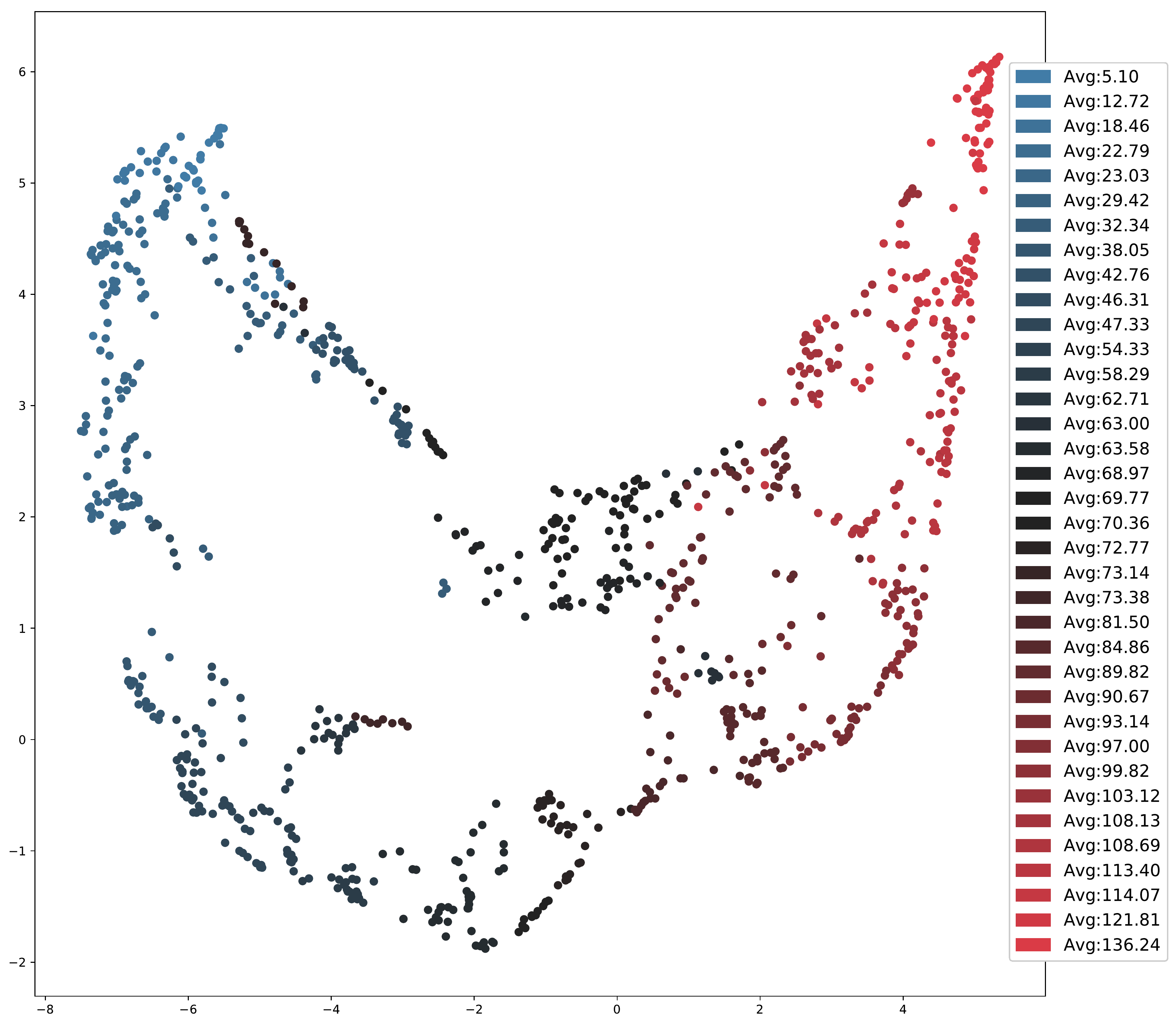}
 	\caption{2D UMAP representation of LSTM encoder activations, where each cluster is colour-coded by the average value of the last timestep across its excerpts. Bluer and redder markers represent time series excerpts that converge to lower- and higher-valued bins at the last timestep, respectively. The colour degradation from left to right suggests that GUM is implicitly learning a quasi-ordering over ordinal bins.}
 	\label{fig:embedding}
\end{figure}

We now provide a visual analysis of the cross-task feature embedding learnt by GUM. We first compute $\mathbf{h}^{(\text{dec})}_0$ (as described in Section \ref{sec:methods}) for a random sample of auxiliary time series excerpts, in addition to all M4 evaluation input sequences. Intra-group minimum variance clusters are then computed through Ward's agglomerative clustering. Results in Figures \ref{fig:frames} and \ref{fig:embedding} are provided for $K=36$ clusters, which yields the largest silhouette score for values between 5 and 50. 

We show the resulting frame clusters in Figure \ref{fig:frames}, observing that those in red are only seen by our GUM network at testing time. We note that the excerpts mainly group towards the right-hand edge of each subplot, i.e. every cluster contains patterns with diverse features (such as degree of smoothness, number of optima and rate of change) that fire similar activations, which then produce similar forecasts at the output layer. 

In Figure \ref{fig:embedding} we show the 2D UMAP \citep{2018arXivUMAP} representation of each frame's feature vector $\mathbf{h}^{(\text{dec})}_0$, colour-coded by cluster id. Bluer and redder markers represent time series clusters that converge to lower- and higher-valued bins at the last timestep, respectively. The smooth colour degradation from left to right, with blue clusters grouping at the left-hand edge and red clusters towards the right, suggests that our model is learning an approximate ordering over ordinal bins.

\subsection{Task 2: Few-shot forecasting}\label{sub:fewshot}
We now assess model performance in the few-shot setting.

\noindent \textbf{Data.} We use the same 20 auxiliary datasets from Section \ref{sub:pred} to train GUM. For the testing phase, we now use 10 series of the monthly-frequency segment of the M4 dataset, those with no missing observations and with at least 1500 measurements available for model inference. All datasets are quantised using $M=150$ bins.

\noindent \textbf{GUM setup.} In this setting, we pre-train a number of GUM models over the auxiliary datasets with different hyperparameters: hidden units $n_h \in \{64, 128, 256, 512\}$, dropout rate $\theta_{\text{drop}}\in\{0.25, 0.5\}$ and $L_2$ regularisation constant $\lambda\in\{10^{-5}, 10^{-6}, 10^{-7},$ $10^{-8}\}$. These serve as initialisation parameters for each M4 time series. Models are trained for up to a further 50 epochs with early stopping and the best hyperparameters for each dataset are chosen by grid search over a validation segment.

\noindent \textbf{Task, baselines and metrics.} We forecast $P_h=100$ measurements ahead from a lookback window of $P=50$, using the same baselines and metrics from Section \ref{sub:pred}, in addition to a naive MOrdReD baseline that has not been pre-trained, i.e. is initialised from scratch and allowed twice as many epochs, also with early stopping. The AR($p$) baseline's lookback is chosen from $p\in \{3, 6, 12, 18\}$.

\noindent \textbf{Results.} We note that our GUM neural network consistently outperforms our baselines across all metrics, as shown in Table \ref{tab:res}. Crucially, GUM also outperforms the naive MOrdReD baseline across all metrics, suggesting that transferring knowledge from prior tasks makes a substantial contribution to predictive performance.

	\section{CONCLUDING REMARKS}
\label{sec:conclusions}
We here introduce a novel neural network model, able to infer probabilistic time series forecasts in both zero- and few-shot learning settings. By introducing a quantisation scheme, we show how knowledge learnt from auxiliary time series can be transferred to previously unseen time series with scarce data. We provide empirical evidence that our approach performs well across a range of metrics, in comparison to other popular methods. By examining the neural activations of our model, we also show how our cross-task feature space encapsulates concepts such as ordinality and knowledge transfer.

	\begin{footnotesize}
	\printbibliography
	\end{footnotesize}

\end{document}